\documentclass[12pt]{article}
\usepackage{amsmath}
\usepackage{bbm}
\usepackage{diagbox}
\usepackage{lineno,hyperref}
\usepackage{subcaption}
\usepackage{graphicx}
\usepackage[style=numeric]{biblatex}
\usepackage{hyperref}

\title{An Efficient Method for the Classification of Croplands in Scarce-Label Regions}
\author{Houtan Ghaffari}
\date{}
\begin{document}
\maketitle
\begin{abstract}
Two of the main challenges for cropland classification by satellite time-series images are insufficient ground-truth data and inaccessibility of high-quality hyperspectral images for under-developed areas. Unlabeled medium-resolution satellite images are abundant, but how to benefit from them is an open question. We will show how to leverage their potential for cropland classification using self-supervised tasks. Self-supervision is an approach where we provide simple training signals for the samples, which are apparent from the data's structure. Hence, they are cheap to acquire and explain a simple concept about the data. We introduce three self-supervised tasks for cropland classification. They reduce epistemic uncertainty, and the resulting model shows superior accuracy in a wide range of settings compared to SVM and Random Forest. Subsequently, we use the self-supervised tasks to perform unsupervised domain adaptation and benefit from the labeled samples in other regions. It is crucial to know what information to transfer to avoid degrading the performance. We show how to automate the information selection and transfer process in cropland classification even when the source and target areas have a very different feature distribution. We improved the model by about 24\% compared to a baseline architecture without any labeled sample in the target domain. Our method is amenable to gradual improvement, works with medium-resolution satellite images, and does not require complicated models. Code and data are available\footnote{\url{https://github.com/houtan-ghaffari/SSCCSLR}}.
\end{abstract}

\section{Introduction}
Although an important affair, many countries do not survey to keep records of the agricultural lands. Access to a complete and up-to-date map of croplands can provide valuable information about agricultural productions, food security, water resources management, greenhouse gas emissions, and the outbreak of some diseases such as avian influenza to relevant organizations and policymakers \cite{Dong2}.

Deep Neural Networks(DNNs) have been used for cropland classification in various forms, and their effectiveness for this task has been proven \cite{8014927}\cite{russwurm2018multi}\cite{pelletier2019temporal}\cite{RUWURM2020421}. However, the main obstacles for cropland classification are not of the model architecture. Working with limited labeled samples is common in remote sensing applications, especially in under-developed areas. On the other hand, deep learning models are infamous to need a large number of labeled samples to achieve acceptable performance. Gathering ample labeled samples requires an enormous effort and might be costly or even impossible to scale up in some situations.

Recently, there has been a surge of interest to make these models as data efficient as possible. A large portion of the proposed algorithms has focused on improving the latent representation of DNNs by using unsupervised methods and make the models prepared for some downstream tasks \cite{hjelm2018learning}\cite{noroozi2016unsupervised}\cite{bojanowski2017unsupervised}. In \cite{Jean_Wang_Samar_Azzari_Lobell_Ermon_2019}, the authors used the word embedding idea from the NLP \cite{mikolov2013efficient}\cite{mikolov2013distributed}, and combined it with a triplet loss \cite{hoffer2018deep} to learn a feature representation from unlabeled high-resolution satellite images.

Using more sophisticated loss functions to provide more control over the latent representation of DNNs is an effective approach to make them data efficient \cite{selflabeling}\cite{caron2020unsupervised}\cite{oord2018representation}. For example, in \cite{8506606}, the authors aimed to learn a metric space by using a neural network for embedding and a loss function based on the distance of samples within a class in that metric space. To alleviate the scarce-label problem, \cite{9254125} proposed a deep multi-view learning method. The authors trained a deep residual network to maximize agreement between differently augmented views of the same scene via a contrastive loss in the latent space. They used this feature representation as input to other machine learning algorithms such as SVM to complete some downstream tasks. 

Transfer learning literature is vast, and there had been significant effort to develop algorithms capable of transferring knowledge from a data-rich source domain into a data-poor target domain. The key idea motivating research in this area was the realization that after humans learn a particular skill, they can perform similar or related tasks better \cite{yang2020transfer}\cite{pan2009survey}. Fine-tuning a pre-trained model is an effective and widely used method in transfer learning when we don't have enough labeled samples in a target domain but the problem is similar to the one in source domain. This method has also been used in remote sensing by pre-training on Image-Net dataset \cite{7342907}. \cite{Xie_Jean_Burke_Lobell_Ermon_2016} improved this idea for a remote sensing application and used night-light prediction as an ancillary task. They trained a fully convolutional model to predict nighttime lights from daytime imagery to learn a proper feature-representation for the few-shot poverty prediction in Africa.

Meta-Learning is another algorithm that has recently become popular for the few-shot learning. The idea behind Meta-Learning is to train on multiple related tasks to learn a function that performs well on a new never-before-seen task after seeing a few examples \cite{schmidhuber1987evolutionary}. For example, finding a good parameter initialization for DNNs is of this kind. In \cite{Alajaji2020FewSS}, the authors tackled the few-shot scene classification using the Model-Agnostic Meta-Learning(MAML) algorithm.

When we aim to tackle global problems such as cropland classification through remote sensing technology, it is important to include the technology's level and facilities in under-developed and developing countries in the equation. Preparing agricultural maps is a cyclic process that needs to be done at least yearly. Hence, the proposed methods must be cheap, efficient, and easy to implement. The previously mentioned works experimented with high-quality hyperspectral satellite images, which are not accessible in most areas and are expensive. It brings difficulties for scientists who want to analyze a scarce-label region through freely accessible satellite images. There is a vast quantity of medium-resolution satellite images, and they are free to access thanks to Landsat and Sentinel projects. In \cite{9150955}, the authors used the MAML algorithm to classify croplands globally. They worked with a coarse-resolution dataset where each sample was a big patch of a Sentinel's image from the Sen12MS dataset \cite{schmitt2019sen12ms}. They acquired the labels from the MODIS Land Cover product MCD12Q1 \cite{Friedl}, which has a coarse-resolution of 500m. It shows a promising line of research toward developing global models. They reported 80\% accuracy on 10-shot.

In this work, our goal is to fill this gap by introducing a new perspective for tackling cropland classification globally. We propose an efficient algorithm for this problem that works well regardless of the neural network's complexity and the quality of satellite images. It does not require a sophisticated model and hyperspectral images, but given that, it is amenable to gradual improvement at multiple levels thanks to its vivid modularity. We will show its efficacy on the few-shot cropland classification and on unsupervised domain adaptation for areas where no labeled data exist.

We have focused on self-supervised methods to improve the cropland classifiers for scarce-label regions as cheaply as possible. Many machine learning algorithms have the self-supervision property one way or another, such as Principal Component Analysis, Independent Component Analysis, or Auto-Encoders. The generalization of this idea as a self-supervised task(also called auxiliary-task and pretext) is a relatively new paradigm in machine learning literature \cite{jing2020self}\cite{anand2019unsupervised}. These methods concern generating artificial labels or guidance from the information in the structure of the data itself. These labels explain some evident properties of the data and are effortless to obtain. The mixture of this elegant idea with DNNs' automatic feature extraction capability has produced an intriguing method for unsupervised representation learning. 

We train a cropland classifier with multiple self-supervised tasks jointly to enrich its feature representation for better performance in few-shot learning. However, it requires these tasks to be well-aligned with each other, or they only consume the model's representation power without helping the main task, which might lead to degraded performance. This alignment concerns the type of information each task extracts from the raw data to improve its performance.

As far as the author knows, there has been one recent work for unsupervised domain adaptation for natural images using a similar approach to what we used for few-shot cropland classification \cite{sun2019unsupervised}. The authors achieved state-of-the-art on four benchmarks. It motivated us to use this method for unsupervised domain adaptation in this problem as well. However, due to limited information in our data, we find it to be harmful here. It is harder to extract cross-geographically shared information from medium-resolution satellite images since they are not as feature-rich as natural images. A crop can reflect different spectral frequencies in distinct regions of the earth because of latent variables such as soil properties and climate, to name a few. Although the tasks are the same, they render a well-performing model in the original domain practically useless in a new environment. Subsequently, we used this idea with another method proposed for domain adaptation that uses a domain detection task \cite{ganin2015unsupervised}. The mixture of two resulted in a more sophisticated algorithm that works well even if the domains have little in common and the data is not feature-rich like natural images. To show the effectiveness of this method, we selected two regions that have a large discrepancy in the statistical distribution of their features.

Our contributions are as follows:
\begin{itemize}
	\item We propose a simple framework to classify croplands, which is especially useful for under-developed areas. This framework organizes the cropland classification problem into separate modules where each one is amenable to gradual improvement thanks to its clean modularity. It is cheap to implement, works well with free medium-resolution images, works well with simple models, and is scalable.
	\item Finding the proper self-supervised tasks for a problem is tricky. We introduce three such tasks compatible with cropland classification that improve the models significantly, whether we have few shots or sufficient data.
	\item Reducing uncertainty is currently of paramount interest in artificial intelligence, and we need to know our algorithms' reliability. We demonstrate that self-supervised tasks reduce the epistemic uncertainty in few-shot learning by always pinpointing a well-performing setting of parameters.
	\item We combine two ideas in the literature for unsupervised domain adaptation and propose a more reliable algorithm that automatically identifies, extracts, and transfers domain-invariant knowledge between domains. It gives us a way to classify unlabeled areas.
\end{itemize}
\section{Methods}\label{sec:Methods}
The proposed methods concern few-shot learning and unsupervised domain adaptation. We compared them to SVM and Random Forest. Also, we show the efficacy of the self-supervised tasks in a standard experiment where we have sufficient labeled samples. SVM and Random Forest are among the most successful and widely used non-deep learning algorithms for cropland classification \cite{Khatami}\cite{rfPelletier}.
\subsection{Model}
\begin{figure}[!t]
	\centering
	\includegraphics[width=\linewidth]{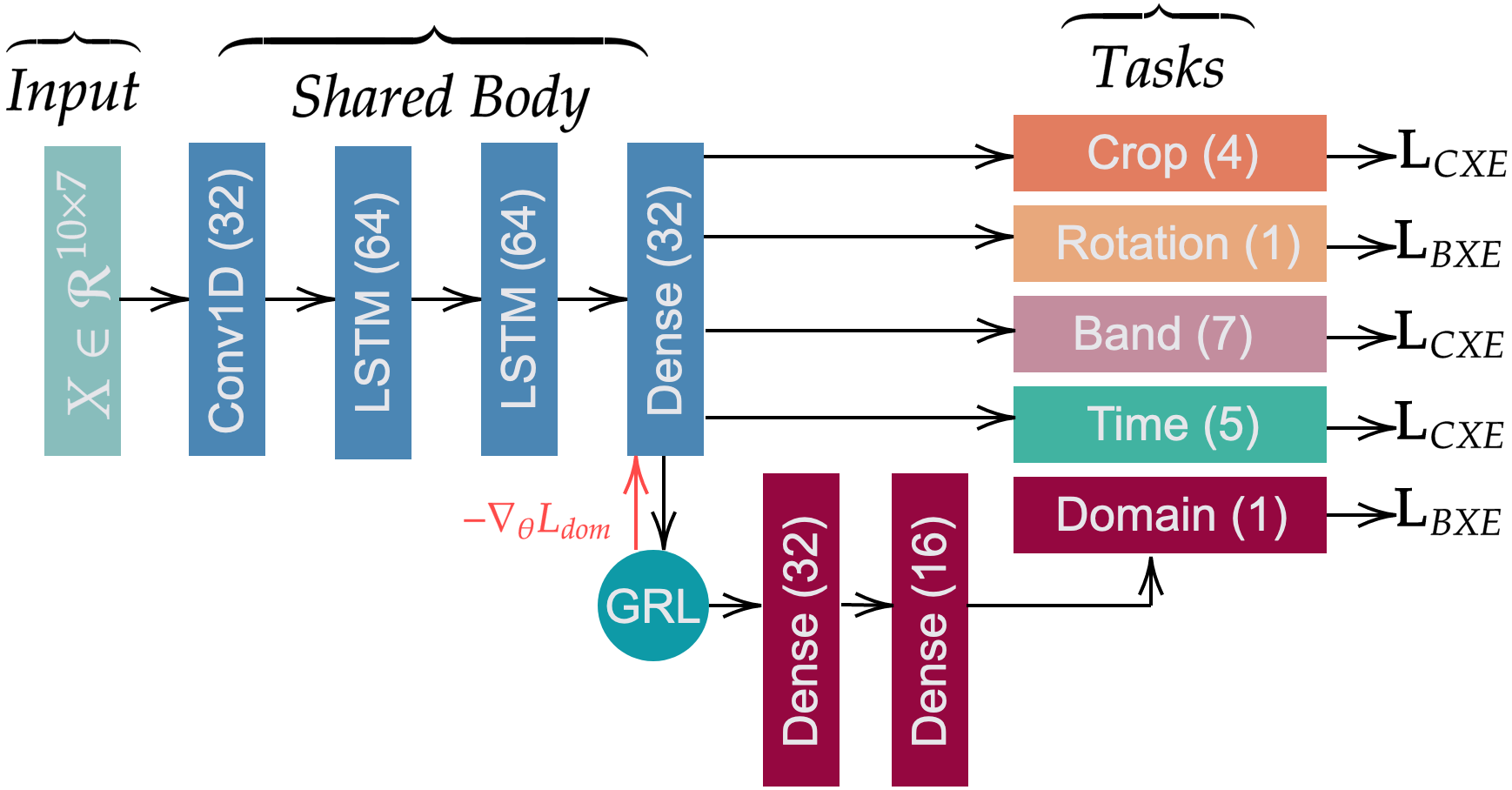}
	\caption{The baseline architecture that we used in all experiments. Tasks are dense layers. Only the main layers are shown. The numbers in parenthesis are the number of units in each layer. $\mathbf{L}_{CXE}$ and $\mathbf{L}_{BXE}$ are categorical and binary cross-entropy loss functions. The layers corresponding to the domain detection task and Gradient Reversal Layer(GRL) are only used in the domain adaptation experiment \cite{ganin2015unsupervised}. In the backpropagation phase, the gradient of the domain classifier's loss gets multiplied by a minus after passing through the GRL module. It means that the shared encoder tries to increase this task's loss by learning indistinguishable features for both domains. In the forward pass, the GRL returns the identity of its input.}
	\label{fig:backbone}
\end{figure}

We have focused on ideas that help us to solve some challenging problems for cropland classification. Hence, we are not concerned with the architecture and hyperparameters. See the model in Fig.\ref{fig:backbone}. In each experiment, we use part of this model or all of it.
\subsection{Data}\label{sec:Data}
\begin{figure}
	\centering
	\includegraphics[width=0.8\linewidth]{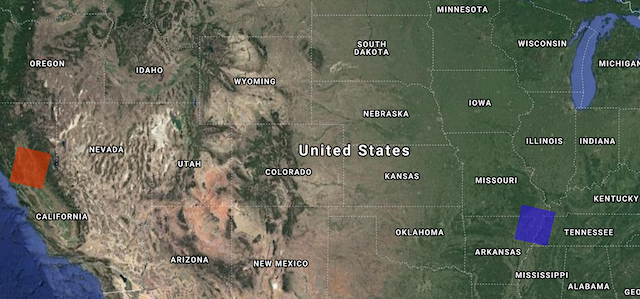}
	\caption{Regions under the study. The red is source, and the blue is target domain.}
	\label{fig:reg_study}
\end{figure}

There are two regions under the study, see Fig.\ref{fig:reg_study}. We used ten Landsat 8 TOA images from the beginning of May to late September in the year 2016. Seven spectral bands were chosen, including Blue, Green, Red, NIR, SWIR 1, SWIR 2, and the Thermal band. We used Google Earth Engine API to download the images \cite{GEE}, and we used Crop Data Layer(CDL) for the ground-truth data \cite{doi:10.1080/10106049.2011.562309}. Four classes from the CDL were selected, including Winter Wheat, Corn, Rice, and Other(everything else). We scaled the images by $10^{-4}$ to bring them in a more suitable range for working with neural networks. It was better than other normalization techniques because it does not change the original trend of the time-series signals, unlike standardization or min-max normalization.

\begin{figure}[t!]
	\centering
	\includegraphics[width=\textwidth]{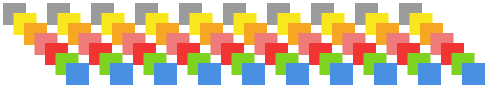}
	\caption[Schematic of one training example]{Schematic of one training example, which consist of one pixel from the time-series of images. Each instance has 10 time-steps and 7 features, which are the spectral bands.}
	\label{fig:input_schematic}
\end{figure}

We used the satellite images in the pixel-wise time-series format. In other words, we only used spectral and temporal information and discarded the spatial information. See the schematic of one input sample in Fig.\ref{fig:input_schematic}. The input space is $\mathcal{X} \subset \mathbf{R}^{10 \times 7}$, and the output space is $\mathcal{Y}=\{0, 1, 2, 3\}$. We show a dataset with $\mathcal{D}=\{(x_i,y_i)| (x_i,y_i) \in \mathcal{X} \times \mathcal{Y}\}_{i=1}^{N}$ where $N$ is the number of samples in the dataset. We indicate the source domain dataset with $\mathcal{D}_S$ and the target domain with $\mathcal{D}_T$. We separated a small dataset from the source domain for few-shot learning, and we refer to it as $\mathcal{D}_{S_{few}}$. See the frequency of each dataset in Tab.\ref{tab:data_frequency}.
\begin{table}[!t]
	\centering
	\caption{Frequency of the datasets.}
	\begin{tabular}{|c|c|c|c|}
		\hline
		\multicolumn{4}{|c|}{Data Frequency}\\ \hline
		& $\mathcal{D}_S$ & $\mathcal{D}_{S_{few}}$  & $\mathcal{D}_T$ \\ \hline
		Wheat & 300000 & 100 & 39000 \\ \hline
		Corn & 300000 & 100 & 300000 \\ \hline
		Rice & 300000 & 100 & 300000 \\ \hline
		Other & 300000 & 100 & 300000 \\ \hline
	\end{tabular}
	\label{tab:data_frequency}
\end{table}
\subsection{Self-Supervised Tasks}\label{sec3:SS_tasks}
The core idea is to extract some labels from the unlabeled data that are both meaningful and easy to obtain. Then, we make self-supervised tasks using these labels. These tasks must align well with each other because they consume the model's representation power. Otherwise, they degrade the main task's accuracy. We have one shared encoder and as many classification layers as we have tasks. Let us call the shared encoder $h(.)$, each classification layer $\mathcal{P}_{t_i}(.)$, and the dataset for each task $\mathcal{D}_{t_i}$. We define the loss function for each task as follows:
\begin{equation}
	L_{t_i}(h, \mathcal{P}_{t_i}, \mathcal{D}_{t_i}) = -\sum_{(x,y) \in \mathcal{D}_{t_i}} y\log\mathcal{P}_{t_i}(y|h(x))
\end{equation}
Now, let us introduce three self-supervised tasks that we find valuable for cropland classification.

\begin{figure}[!t]
	\centering
	\includegraphics[width=0.8\linewidth]{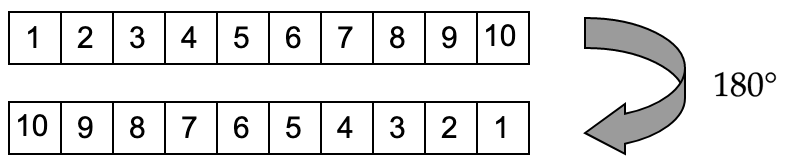}
	\caption{This schematic shows the rotation detection task. The array at the top depicts a sample's time axis ordered relative to the times of observation. We rotate it on the time axis to sort the observation dates in reverse order, shown in the bottom array. We label the regular sample with zero and the rotated version with one. We are not showing the feature axis here.}
	\label{fig:rotation_task}
\end{figure}

\textbf{Rotation detection}:\hspace{1em}To make this task, we make a copy of the original dataset and invert all the samples on their time axis. In other words, we reverse the order of observations from the last date to the first one. Then, we assign the label of zero to the original samples and one to the rotated samples, which means not-rotated and rotated. Let us name the dataset for this task $\mathcal{D}_R$. See Fig.\ref{fig:rotation_task} for an intuitive picture of generating this dataset. The model has to learn if a sample is rotated or not. The auxiliary output layer is a dense layer with one unit and the sigmoid activation function since this is a binary task. We use binary cross-entropy for the loss function. Let us indicate this classification layer with $P_R(.)$. The loss function for this task is defined as follows:
\begin{equation}
	L_{R}(h, \mathcal{P}_{R}, \mathcal{D}_{R}) = -\sum_{(x,y) \in \mathcal{D}_{R}} y\log\mathcal{P}_{R}(y|h(x))
\end{equation}

\begin{figure}[!t]
	\centering
	\includegraphics[width=0.8\linewidth]{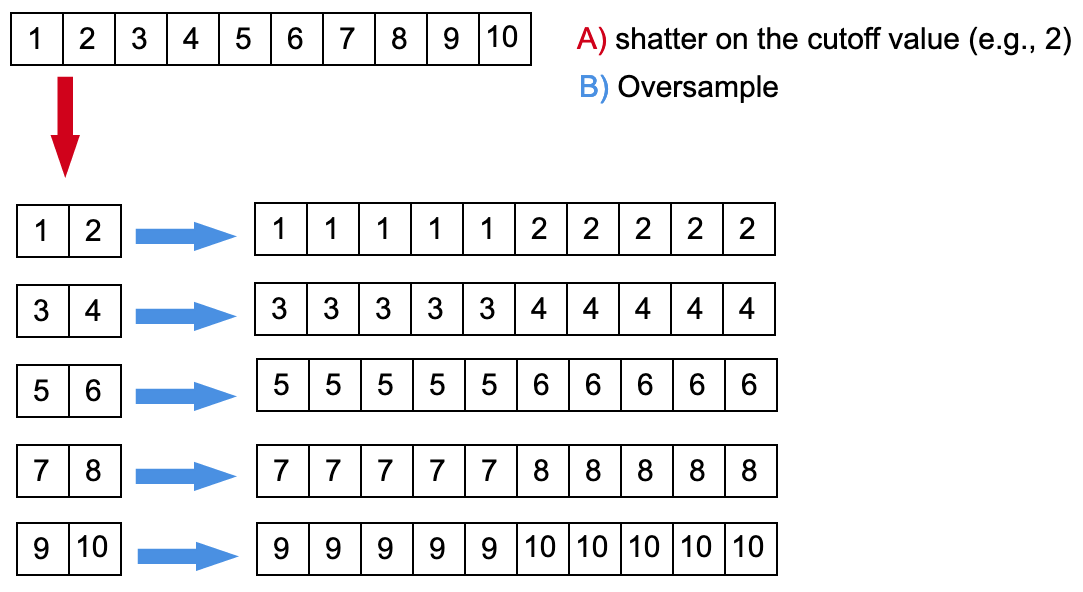}
	\caption{This schematic shows the time-segment detection task. Choose a cutoff value to shatter a sample into equally sized segments on the time axis. Copy each time-segment on the time axis to match its shape with a regular training example. Make sure to keep the dates of observation in the correct order. Then, Give each time-segment a unique label. We are not showing the feature axis here.}
	\label{fig:time_seg_task}
\end{figure}

\textbf{Time-Segment detection}:\hspace{1em}To make this task, first, we need to choose a cutoff point on the time axis. It's a hyperparameter, and depending on the number of observations in the time-series, we can vary the cutoff value. Our dataset consists of ten time-steps, and we chose the value of two for the cutoff. Hence, each regular sample $x \in \mathcal{R}^{10 \times 7}$ breaks into five time-segments $\hat x_j \in \mathcal{R}^{2\times7}$ for $j=0,...,4$. After this shattering, we copy each $\hat x_j$ on the time axis, without mixing the order of observations, so their shape match with a regular sample $x \in \mathcal{R}^{10\times7}$. We assign labels from zero to the number of time-segments to each one(e.g., 0 to 4 in our case). i.e., $\hat x_0$ gets the label of zero, $\hat x_1$ gets the label of one, and so on. Let us name the dataset for this task $\mathcal{D}_{TS}$. See Fig.\ref{fig:time_seg_task} for an intuitive picture of generating this dataset. The model has to detect the time-segment of each sample. The auxiliary output layer is a dense layer with five units and the softmax activation function since this a multi-class task. We use categorical cross-entropy for the loss function. Let us indicate this classification layer with $P_{TS}(.)$. The loss function for this task is defined as follows:
\begin{equation}
	L_{TS}(h, \mathcal{P}_{TS}, \mathcal{D}_{TS}) = -\sum_{(x,y) \in \mathcal{D}_{TS}} y\log\mathcal{P}_{TS}(y|h(x))
\end{equation}

\begin{figure}[!t]
	\centering
	\includegraphics[width=0.8\linewidth]{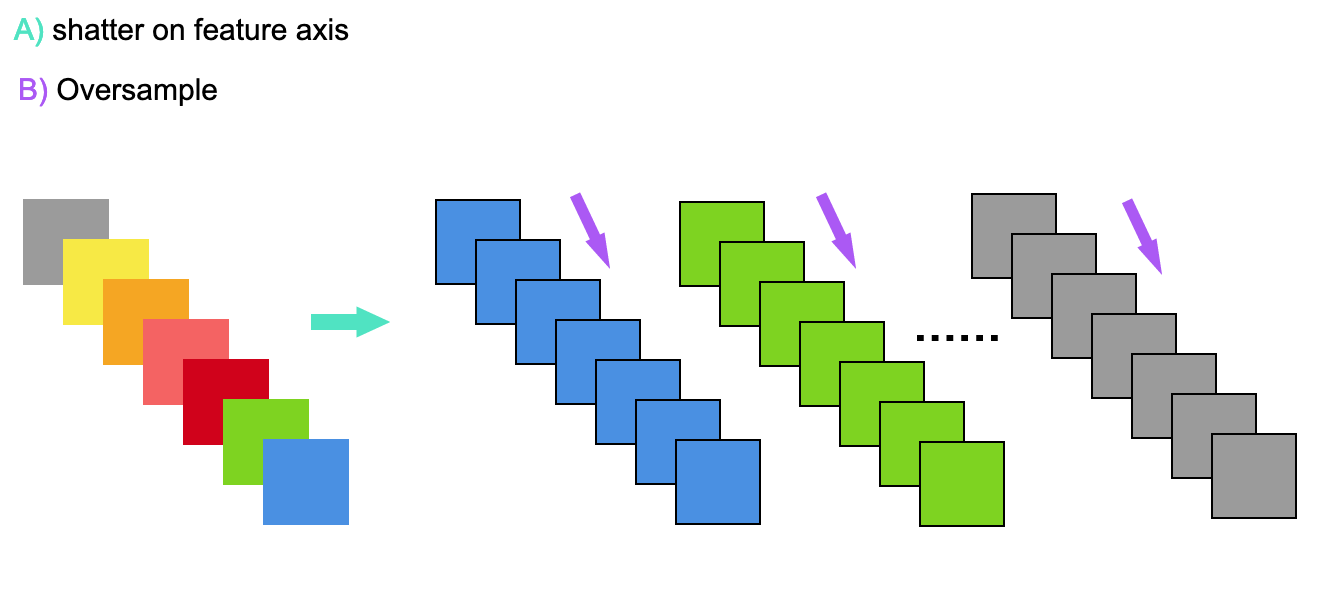}
	\caption{This schematic shows the band detection task. We shatter each instance on the feature axis, which is spectral bands here. Afterward, we copy each isolated spectral band on the feature axis to match its shape with a regular input. Finally, we give a unique label to each separated time-series of spectral bands. We are not showing the time axis here.}
	\label{fig:band_task}
\end{figure}

\textbf{Band detection}:\hspace{1em}Here, the model has to detect the spectral bands. Our dataset consists of seven spectral bands. We separate each spectral band from the samples, which results in seven shards $\hat x_b \in \mathcal{R}^{10\times1}$ for $b=0,...,6$, from each original $x \in \mathcal{R}^{10\times7}$. Afterward, we have to oversample each $\hat x_b$ on the feature axis to match its shape with a regular one. Notice that we have a feature at each time step, and each one gets copied separately to end up with $\hat x_b \in \mathcal{R}^{10\times7}$ for $b=0,...,6$. We give each spectral band a label from zero to six. Let us name the dataset for this task $\mathcal{D}_B$. See Fig.\ref{fig:band_task} for an intuitive picture of generating this dataset. The auxiliary output layer is a dense layer with seven units and the softmax activation function, and categorical cross-entropy is the loss function. Let us indicate this classification layer with $P_B(.)$. The loss function for this task is defined as follows:
\begin{equation}
	L_{B}(h, \mathcal{P}_{B}, \mathcal{D}_{B}) = -\sum_{(x,y) \in \mathcal{D}_{B}} y\log\mathcal{P}_{B}(y|h(x))
\end{equation}
\subsection{Few-Shot Learnering by Self-Supervision}\label{sec3:fewshot_SS}
We have a few labeled samples in a region and many unlabeled ones. We want to improve the model by leveraging the information that resides in the unlabeled samples through the self-supervised tasks that we just defined. We use $\mathcal{D}_S$ to make the self-supervised tasks. It is also our test set, and $\mathcal{D}_{S_{few}}$ is our training set for the cropland classification task. Its output layer is a dense layer with four units and the softmax activation function. We use the categorical cross-entropy loss function since we have four classes. Let us indicate this classification layer with $P_C(.)$. The loss function for this task is defined as follows.
\begin{equation}
	L_{C}(h, \mathcal{P}_{C}, \mathcal{D}_{S_{few}}) = -\sum_{(x,y) \in \mathcal{D}_{S_{few}}} y\log\mathcal{P}_{C}(y|h(x))
\end{equation}
Hence, our overall cost function is defined as follows:
\begin{equation}
	\mathcal{L}_{total} = L_C + L_R + L_{TS} + L_B
\end{equation}
The hope is that the self-supervised tasks improve the model's encoder by enriching its feature representation since the main task has only a few labeled samples. We optimize this cost function jointly while each classification head is responsible for its own dataset. The model that we used for this experiment is the one depicted in Fig.\ref{fig:backbone} without the modules for the domain detection task. We did not weigh the loss functions and find it unnecessary. It was said to be unnecessary by \cite{sun2019unsupervised} too.
\subsection{Domain Adaptation by Self-Supervision}\label{sec3:DA}
Unsupervised domain adaptation is a type of transfer learning where we have labeled samples in the source domain and unlabeled samples in the target domain \cite{pan2009survey}. Let $\mathcal{X}_S$ indicates the feature space for the source domain and $\mathcal{X}_T$ for the target domain. We show the underlying distribution of features with $\mathcal{P}_S(\mathcal{X}_S)$ for the source domain and $\mathcal{P}_T(\mathcal{X}_T)$ for the target domain. The problem here is that we have $\mathcal{X}_S = \mathcal{X}_T$ but $\mathcal{P}_S(\mathcal{X}_S) \neq \mathcal{P}_T(\mathcal{X}_T)$.

\begin{figure}[t!]
	\centering
	\begin{subfigure}{0.7\textwidth}
		\centering
		\includegraphics[width=\textwidth]{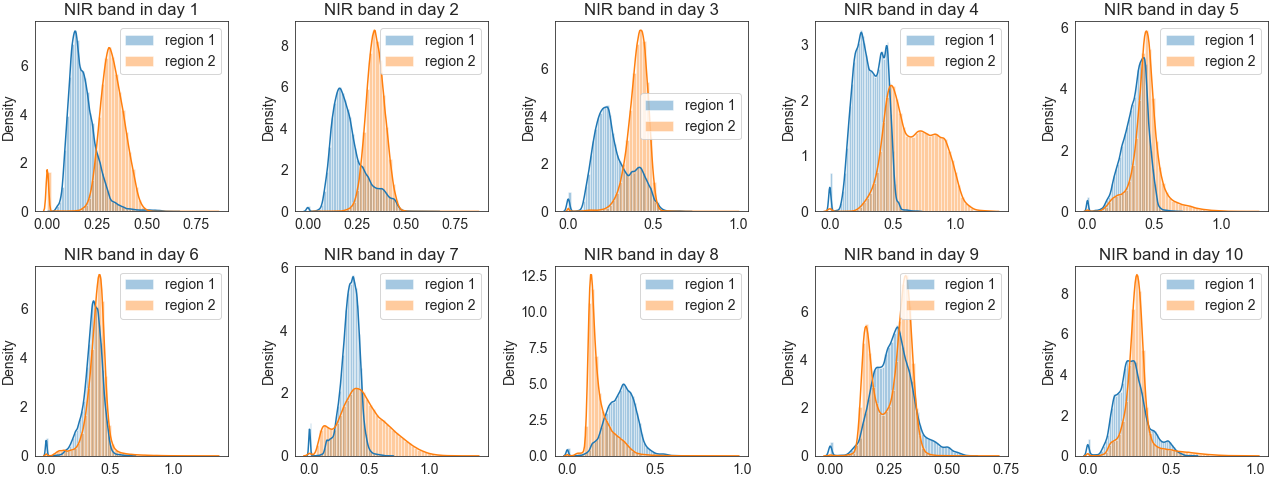}
		\caption{}
	\end{subfigure}
	\hfill
	\begin{subfigure}{0.25\textwidth}
		\centering
		\includegraphics[width=\textwidth]{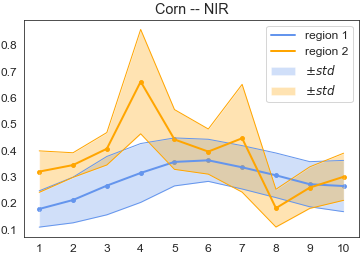}
		\caption{}
	\end{subfigure}
	\caption{(a) Empirical distribution of NIR, (b) The mean of NIR with one standard deviation for the corn class in 10 days of observation. The orange color is for the target and the blue is for the source domain.}
	\label{fig:EPMF_corn_nir}
\end{figure}

We want to encourage the model to finds a shared feature representation for both domains. \cite{sun2019unsupervised}, proposed to train the self-supervised tasks jointly with the main one, just like the way we used for few-shot learning. However, the unlabeled dataset consists of both the target and source domain samples. Therefore, the self-supervised tasks push the feature representation of both domains into a shared region of the latent space. They achieved state-of-the-art results on four benchmarks for natural images. It didn't work for us. The reason probably is that our domains differ too much, and our satellite images are not as feature-rich as natural images. We calculated the empirical distribution of the spectral bands in each region to see their differences. For the sake of brevity, here we only show the NIR band for the corn class, see Fig.\ref{fig:EPMF_corn_nir}. There is a significant difference in the statistical distribution of features between the domains.

Subsequently, we used an additional ancillary task named domain detection, which is particularly useful for domain adaptation and needs to be combined with a component called the Gradient Reversal Layer(GRL) \cite{ganin2015unsupervised}. The marriage of these two ideas resulted in a method that can pinpoint useful information automatically by using self-supervised tasks. Moreover, thanks to the domain detection task, it keeps only the relevant knowledge between the domains as much as possible and does not overfit on the source domain for the main task, which does not have any label in the target domain. Hence, we can worry less about the negative transfer even when our regions differ a lot.

We combine $X_S \in \mathcal{D}_S$ and $X_T \in \mathcal{D}_T$ into one unlabeled dataset $\mathcal{D}_U$. Then, we make our three self-supervised tasks using $\mathcal{D}_U$. It's quite simple to produce a dataset for the domain detection task. We have to label the samples from $\mathcal{D}_S$ with zero and the samples from $\mathcal{D}_T$ with one. Then, we append a small module to the shared encoder for discriminating between the samples drawn from each domain. However, there is a big problem here. This task makes the feature representation more diverse, while we want it to become unified for different domains. That is why we use the GRL. See Fig.\ref{fig:backbone} for an intuitive understanding of what GRL does. Its role is to encourage the encoder to finds a feature representation that makes the domain detector's job as hard as possible. Therefore, the encoder has to learn an indistinguishable feature representation for both domains.
\section{Results}\label{sec:Results}
In this section, we report the results of our experiments. First, we demonstrate how self-supervised tasks help in the few-shot scenario. Then, we show their effectiveness for unsupervised domain adaptation for cropland classification, where there is no labeled sample in the target domain.
\subsection{Few-Shot Learning Results}
Self-supervised tasks were examined and tested individually or in combination to aid the main task. There is no validation set in this experiment to guide the training and adjust the model's hyperparameters. Hence, there is a big problem with how many epochs to train the model. We shall see that the self-supervised tasks can reduce this concern by increasing the model's convergence rate.

We used $\mathcal{D}_{S_{few}}$ for the training set and $\mathcal{D}_S$ for the test set. We have trained our models for 12000 epochs on the main task's few-shots. For the multi-task models with one self-supervised task, we trained on the self-supervised dataset for 20 epochs, and for the one with all three tasks, we trained on the self-supervised dataset for 10 epochs. We used the Adam optimization algorithm with a constant learning rate of 0.001 \cite{kingma2014adam}, and we conducted each experiment 30 times. See the mean accuracies with one standard deviation in Tab.\ref{tab:SS_acc}. The self-supervised tasks make the model robust against noise and increase its convergence rate. Consistent convergence to well-performing parameters indicates that this method reduces epistemic uncertainty. In few-shot learning, we need the model to show a stable performance to use it with more confidence in a real-world application, which is the case in our results. The model becomes less sensitive to parameter initialization and regularization methods.
\begin{table}[t!]
	\centering
	\caption[Accuracies in few-shot learning]{Baseline is our model only with the cropland classification task. R stands for the Rotation, T stands for the Time-Segment, and B stands for the Band detection task. We ran each experiment 30 times and reported the mean accuracy with one standard deviation. We have run SVM and Random Forest only once.}
	\resizebox{\columnwidth}{!}{%
		\begin{tabular}{|c|c|c|c|c|c|}
			\hline
			\multicolumn{6}{|c|}{Accuracy (\%)}\\ \hline
			\diagbox[dir=NW]{Model}{Experiment} & $5$-shot        & $10$-shot       & $20$-shot       & $50$-shot       & $100$-shot \\ \hline
			Baseline            & $35.19\pm7.13$          & $24.94\pm0.09$          & $25.62\pm1.8$          & $88.22\pm0.5$          & $92.07\pm0.28$ \\ \hline
			Baseline + R         & $81.21\pm4.04$ & $83.34\pm3.22$          & $85.84\pm1.11$ & $91.82\pm0.42$          & $93.57\pm0.42$ \\ \hline
			Baseline + T         & $82.58\pm2.48$          & \textbf{87.49$\pm$1.13} & \textbf{88.86$\pm$2.94}         &  \textbf{92.33$\pm$0.8}          & \textbf{94.02$\pm$0.39} \\ \hline
			Baseline+ B        & \textbf{82.84$\pm$1.31}         & $83.62\pm0.76$          & $86.93\pm0.78$          & $90.20\pm0.48$          & $92.16\pm0.26$ \\ \hline
			Baseline + T + B + R & $81.83\pm4.51$          & $85.01\pm1.58$          & $88.51\pm1.14$          &$91.76\pm0.63$ & $93.73\pm0.48$ \\ \hline
			SVM & $80.49$ & $80.20$ & $83.40$ & $86.59$ & $88.81$ \\ \hline
			Random Forest & $79.82$ & $81.74$ & $84.45$ & $89.03$ & $91.69$ \\ \hline
		\end{tabular}
	}
	\label{tab:SS_acc}
\end{table}
\subsection{Do Self-Supervised Tasks Increase the Accuracy?}
There is an important question that we can't answer with certainty based on few-shot results. The self-supervised tasks increased the convergence rate of the model significantly. But, do they also bring useful and new information? In other words, if we had enough data to not worry about the convergence and stability of the model, is there any reason to still use them? The answer is positive. We will show that self-supervised tasks can also increase accuracy.
\begin{table}[t!]
	\centering
	\caption[Accuracies in self-supervised learning for increasing the accuracy]{Standard experiment to show the capability of the self-supervised tasks in increasing the accuracy of the baseline model.}
	\begin{tabular}{|c|c|c|c|c|c|}
		\hline
		\multicolumn{2}{|c|}{Accuracy (\%)}\\ \hline
		Baseline  & $96.63$ \\ \hline
		Baseline + R & $98.08$ \\ \hline
		Baseline + T & \textbf{98.1} \\ \hline
		Baseline + B & $96.83$ \\ \hline
		Baseline + T + B + R & $98.07$  \\ \hline
		SVM & 97.48 \\ \hline
		Random Forest & $97.54$ \\ \hline
	\end{tabular}
	\label{tab:SS_acc_normal}
\end{table}
\begin{figure}[t!]
	\centering
	\begin{subfigure}{0.19\textwidth}
		\centering
		\includegraphics[width=\textwidth]{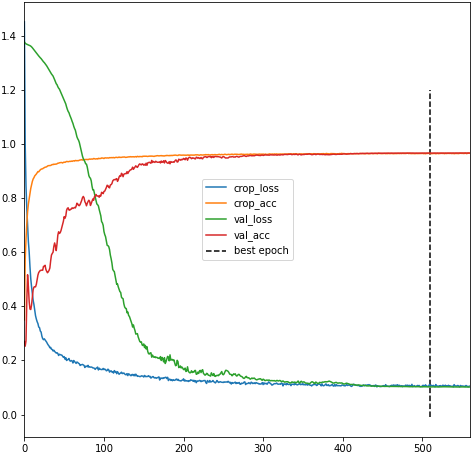}
		\caption{Baseline.}
	\end{subfigure}
	\hfill
	\begin{subfigure}{0.19\textwidth}
		\centering
		\includegraphics[width=\textwidth]{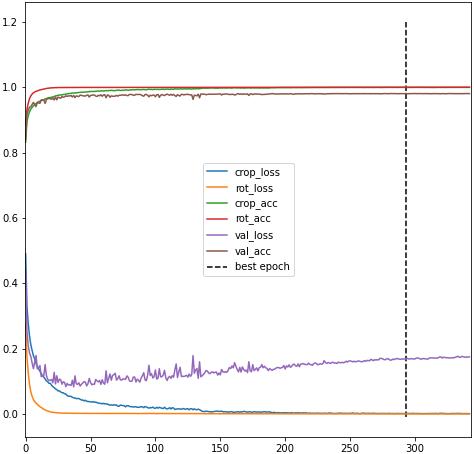}
		\caption{Rotation.}
	\end{subfigure}
	\hfill
	\begin{subfigure}{0.19\textwidth}
		\centering
		\includegraphics[width=\textwidth]{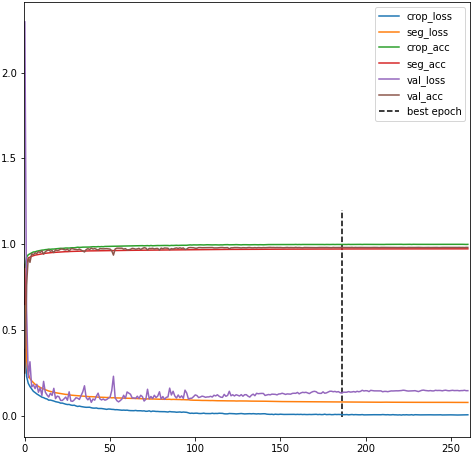}
		\caption{Time-Seg.}
	\end{subfigure}
	\hfill
	\begin{subfigure}{0.19\textwidth}
		\centering
		\includegraphics[width=\textwidth]{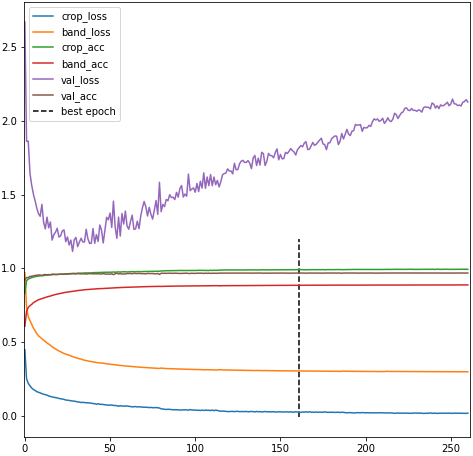}
		\caption{Band.}
	\end{subfigure}
	\begin{subfigure}{0.19\textwidth}
		\centering
		\includegraphics[width=\textwidth]{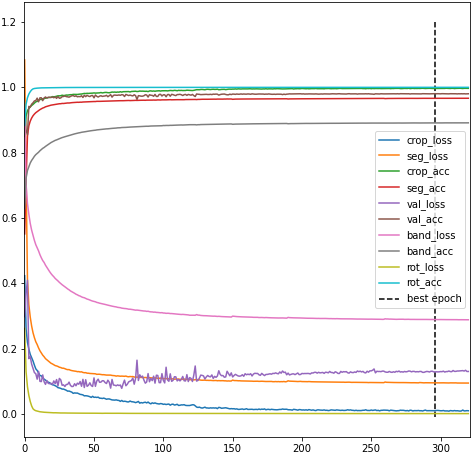}
		\caption{Three tasks.}
	\end{subfigure}
	\hfill
	\caption[Training progress in standard experiment using auxiliary tasks]{Training history of the models in standard experiment.}
	\label{fig:progress_normal}
\end{figure}

For this experiment, We divided the $\mathcal{D}_S$ into a training set including 9000, a validation set including 21000, and a test set including 270000 samples from each class. Afterward, we trained all of the models for 1000 epochs with patience of 25 epochs. After every 25 epochs, if the accuracy did not improve on the validation set, we halved the learning rate. We set the minimum learning rate to 0.00005, and if we did not observe an improvement with this learning after 25 epochs, we halt the training and restore the parameters based on the best epoch on the validation set. See the results in Tab.\ref{tab:SS_acc_normal}. We achieved a 1.5\% boost compared to the baseline(no self-supervised task) even in this tight margin of error. Also, we recorded the training history for each model, see Fig.\ref{fig:progress_normal}. The training histories also indicate how significantly the self-supervised tasks increased the convergence rate, while the baseline took about 300 epochs to converge(on the validation set accuracy). Perceive that self-supervised tasks do not cause overfitting if we keep training the model, which is another good sign of their compatibility with the main task. Another benefit is that we can train our models faster than the baseline, even if one epoch might take longer. We only need 20 to 30 epochs to converge. For SVM and Random Forest, we performed a grid search on hyperparameters and selected the best models based on validation set accuracy. Please notice that, we used a very simple architecture for our experiments. Given a more sophisticated model, we could probably achieve better results without worrying too much for convergence or overfitting due to a large number of parameters.
\subsection{Domain Adaptation Results}\label{sec4:DA_results}
For the main task, we used all the labeled samples from $\mathcal{D}_S$, and we test the models on $\mathcal{D}_T$. We made the self-supervised tasks from the union of all samples from both domains(see section \ref{sec3:DA}). Notice that we did not use the crop labels from $\mathcal{D}_T$ for the training. For optimization, we used the Adam algorithm with a constant learning rate of 0.001. We trained the models for 300 epochs, and we compared the performance with Random Forest and a baseline, both of which were trained just with the source domain labeled samples for cropland classification. SVM does not scale well since we are using 1.2M samples from the source domain. Hence, it has been omitted from this experiment. See Tab.\ref{tab:acc_DA} for the results.
\begin{table}[t!]
	\centering
	\caption{Results of the domain adaptation experiment. The first row shows a simple model that was trained only with source domain labeled samples. The second row belongs to a model that was enhanced with self-supervised tasks without domain detection. The third row is the baseline model enhanced only with the domain detection task. The fourth row shows the model that was enhanced with all of the tasks. Capital letters are the first letter in each task's name.}
	\begin{tabular}{|c|c|}
		\hline
		\multicolumn{2}{|c|}{Accuracy $(\%)$} \\ \hline
		& Target Domain \\ \hline
		Baseline & 47.66 \\ \hline
		Baseline + R + T + B & 45.56 \\ \hline
		Baseline + D & 51.8 \\ \hline
		Baseline + R + T + B + D & 71.06 \\ \hline
		Random Forest & 51.17 \\ \hline
	\end{tabular}
	\label{tab:acc_DA}
\end{table}

It's surprising how self-supervised tasks alone can't help us with this problem(Compare the 1st and 2nd rows in Tab.\ref{tab:acc_DA}). It is in contradiction with natural images \cite{sun2019unsupervised}. The reason is that our source and target domains have quite different distributions. Also, satellite images at the temporal, spatial, and spectral resolution that we used are not rich enough to find shared regularities in such diverse domains easily. After adding the domain detection task, the conclusion changed(4th row in table \ref{tab:acc_DA}). It shows the importance of the domain detection task for efficient domain adaptation \cite{ganin2015unsupervised}. Self-supervised tasks helped the model to become more domain-invariant when combined with this new task. However, they degraded the main task's performance when used alone because the regions were not close enough. If we take the overfitting in a more abstract sense, the model is overfitting on the source region for cropland classification, but the domain detection task prevents it and makes all the self-supervised tasks retain the common information between both regions as much as possible. Therefore, the main task has very little room for overfitting on the source domain. It is an efficient method to find, extract, and transfer the most useful information automatically from the source domain, and it is easy to improve this framework by introducing more relevant tasks for cropland classification and using more sophisticated models. It makes us worry less about selecting similar regions, and we can use all of our labeled samples more efficiently. However, if we have similar regions then it's better to use them or at least weigh their samples higher.
\section{Conclusion}\label{sec:Conclusion}
We presented a framework based on self-supervision for cropland classification. It is effective for both the few-shot and unsupervised cropland classification(given a labeled source domain). It works well with free medium-resolution satellite images, although the methodology is valid regardless of images' quality. Moreover, it is model-agnostic in the sense that it can boost almost any neural network. We showed its efficacy by using one of the most simple architectures possible. It is a cheap and proper approach for developing reasonably good models in regions where ground-truth data is scarce. We introduced three compatible self-supervised tasks for cropland classification. In the few-shot setting, we showed they increased the convergence rate, increased the accuracy, and reduced epistemic uncertainty by always converging the model to well-performing parameters. Also, we demonstrated that they bring these benefits along even if we have enough labeled samples. Subsequently, we combined these tasks with a domain detection task, which allowed us to perform unsupervised domain adaptation successfully. This combination works even if the regions have a large discrepancy, like in our case. It automatically identifies, extracts, and transfers the essential knowledge from the source geographic region. More importantly, this method is amenable to improvement at multiple levels and is particularly suited for under-developed areas. One can construct more self-supervised tasks compatible with cropland classification. Also, more sophisticated neural networks have already shown great success for this problem. Additionally, taking spatial information into account opens up new possibilities for both the architecture and self-supervised tasks.
\section*{Acknowledgment}
I would like to thank Marc Ru{\ss}wurm from the Technical University of Munich for his support and fruitful discussions, which improved this work considerably.

\printbibliography

\end{document}